\documentclass[letterpaper, 10 pt, conference]{ieeeconf}  

\IEEEoverridecommandlockouts                              

\usepackage{graphics} 
\usepackage{graphicx}
\usepackage{subcaption}
\usepackage{amsmath} 
\usepackage{amssymb}  
\usepackage{multirow}
\title{\LARGE \bf
OpenFusion++: An Open-vocabulary Real-time Scene Understanding System
}

\author{Xiaofeng Jin$^{*1}$, Matteo Frosi$^{1}$ and Matteo Matteucci$^{1*}$
\thanks{$^{*}$Corresponding author.}
\thanks{$^{1}$All authors are affiliated with Politecnico di Milano, Milan 20133, Italy. {\tt\small \{xiaofeng.jin, matteo.frosi, matteo.matteucci\}@polimi.it}}%
}%

\begin{document}

\maketitle
\thispagestyle{empty}
\pagestyle{empty}

\begin{abstract}

Real-time open-vocabulary scene understanding is essential for efficient 3D perception in applications such as vision-language navigation, embodied intelligence, and augmented reality. However, existing methods suffer from imprecise instance segmentation, static semantic updates, and limited handling of complex queries. To address these issues, we present OpenFusion++, a TSDF-based real-time 3D semantic-geometric reconstruction system. Our approach refines 3D point clouds by fusing confidence maps from foundational models, dynamically updates global semantic labels via an adaptive cache based on instance area, and employs a dual-path encoding framework that integrates object attributes with environmental context for precise query responses. Experiments on the ICL, Replica, ScanNet, and ScanNet++ datasets demonstrate that OpenFusion++ significantly outperforms the baseline in both semantic accuracy and query responsiveness.

\end{abstract}

\section{Introduction}

Open-vocabulary scene understanding using RGB-D data is crucial for embodied intelligence, autonomous navigation, augmented reality (AR), and robotics applications. Unlike closed-set recognition systems that depend on predefined categories, open-vocabulary approaches utilize Vision-Language Foundation Models (VLFMs), such as CLIP~\cite{4} and its variants~\cite{1}, to enable systems to recognize and reason about previously unseen concepts.

In recent years, open-vocabulary 3D scene understanding has made notable strides by combining category-agnostic segmentation models (e.g., Segment Anything Model, SAM~\cite{2}) with multi-modal alignment frameworks (e.g., CLIP). State-of-the-art systems in the literature share core elements in their pipelines. First, they employ SAM to generate initial object masks, then use multi-scale CLIP feature extraction for region-level semantic embedding projection, and finally leverage depth sensors and multi-view fusion to refine 3D semantic reconstruction. However, this two-stage pipeline suffers an inherent drawback: the class-agnostic 2D segmentation models' tendency to over-segment produces numerous redundant masks, expecting the system to compute features independently for each mask and severely hindering real-time performance.

Combining robotics with VLFMs enables higher-level decision-making, which requires attention to scalability and real-time performance. OpenFusion~\cite{3} addressed this by using a TSDF-based method for efficient geometric mapping and introducing SEEM~\cite{24} to reference masks with object semantics and embeddings directly. This design significantly improves real-time responsiveness and scalability by eliminating the traditional mask post-processing pipeline.

During real-time scene exploration, occlusions and geometric variations challenge SEEM-based segmentation in maintaining consistent semantic boundaries, leading to semantic confusion (e.g., sofa versus pillow) and semantic drift due to partial observations. Additionally, OpenFusion uses a static semantic initialization strategy, relying on the first observed semantic embedding as a global descriptor, which overlooks the refinement potential of incremental observations and causes semantic ambiguities to accumulate. More critically, SEEM’s closed-form attribute regression fails to capture contextual relationships between instances and their environment (e.g., “a table near the window”), limiting the system's ability to handle complex queries.

To address these challenges, we present OpenFusion++, an open-vocabulary real-time scene understanding system. Based on OpenFusion’s TSDF-based architecture, our approach enhances scene understanding by four contributions.
\begin{itemize}

\item We propose a confidence-guided 3D point sampling strategy. By incorporating pixel-level uncertainty estimations from foundation models (e.g., SEEM), we aim to resolve instance boundary semantic conflicts.

\item We propose an adaptive semantic caching mechanism, which dynamically updates instance embeddings based on the observed physical area, mitigating semantic drift.

\item We propose a dual-branch query framework that integrates object-centric features (from SEEM) and environment-aware embeddings (from a CLIP variant) to accurately respond to nested queries of similar instances through a hierarchical retrieval mechanism.

\item We conduct quantitative and qualitative evaluations on multiple sequences of various datasets, including ablation studies on the various contributions of the system.
\end{itemize}

\begin{figure*}[thpb]
  \centering
  \includegraphics[width=1.0\textwidth]{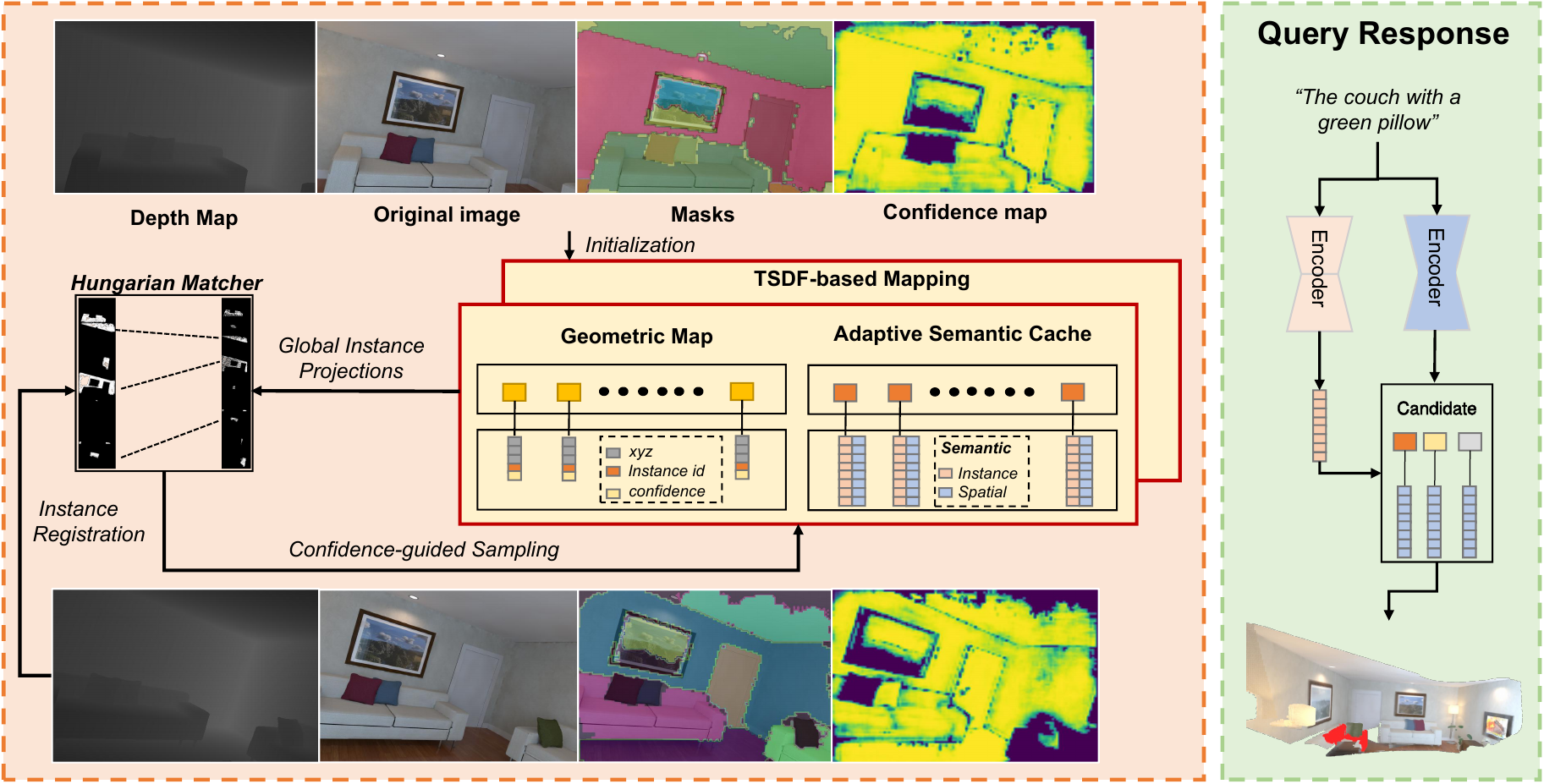}
  \caption{Overview of OpenFusion++ and its three core modules. 1) Real-time TSDF-based geometric reconstruction that dynamically fuses multi-view RGB-D data into sparse voxel blocks. 2) Dual-structure map consisting of a management-geometric map that organizes voxel blocks, and a semantic map that stores instance-level features (SEEM object semantics and CLIP spatial context) via adaptive caching. 3) Hierarchical query architecture that integrates object attributes and environmental features through a two-stage retrieval (coarse filtering and fine-grained matching) to resolve semantic queries.}
  \label{overview}
\end{figure*}

\section{Related Works}

\subsection{Open-Vocabulary 2D Segmentation}
Recent advancements in vision-language foundation models (VLFMs) have significantly improved open-vocabulary segmentation. CLIP~\cite{4} pioneered the use of large-scale image–text contrastive learning to align visual features with textual descriptions, enabling recognition beyond predefined categories. DINO~\cite{5, 6} and MAE~\cite{7} further enhanced feature generalization through self-supervised learning, while GroupViT~\cite{8} and SegCLIP~\cite{9} integrated multimodal contrastive training to improve segmentation quality.

For instance-level segmentation, category-agnostic models such as SAM~\cite{3} and SEEM~\cite{24} have emerged as key components for open-world perception. These methods generate instance masks without relying on predefined labels, making them well-suited for open-vocabulary tasks. However, SAM tends to over-segment objects, leading to redundant masks and increased computational overhead. Meanwhile, SEEM references object masks with explicit semantic embeddings but struggles with instance boundary inconsistencies under occlusions or viewpoint variations

\subsection{Open-Vocabulary 3D Scene Understanding}
Extending open-vocabulary recognition to 3D environments requires integrating 2D segmentation with spatial mapping. ConceptFusion~\cite{10} combined SAM with CLIP-based embeddings to achieve region-level 3D comprehension, while OpenMask3D~\cite{11} refined instance-level queries through multi-scale feature fusion. OpenSU3D~\cite{12} introduced GPT-4V for detailed instance descriptions, enabling improved scene-level reasoning. Open3DIS~\cite{13} tackled object diversity by leveraging 2D mask guidance for 3D instance segmentation, and OpenIns3D~\cite{14} explored 3D instance retrieval without relying on image inputs.

Despite these advances, many approaches remain sensitive to occlusions and struggle to maintain consistent instance semantics across views. OpenScene~\cite{15} aligned 3D point clouds with CLIP features for cross-modal queries (e.g., “soft areas”) but lacked instance-level segmentation. Similarly, SAM3D~\cite{16} and MaskClustering~\cite{17} introduced view-consensus clustering but faced label ambiguities due to limited global semantic updates.

\subsection{Real-Time Scene Understanding}

For robotic perception, real-time scene understanding requires efficient mapping and incremental semantic updates. ESAM~\cite{18} employed lightweight segmentation models such as FastSAM~\cite{19} to accelerate instance extraction, while PanoSLAM~\cite{21} incorporated Gaussian splatting~\cite{22} for joint panoptic-semantic modeling. OVO-SLAM \cite{23} combined instance fragment tracking with multi-view CLIP fusion to support query-based scene interpretation.

OpenFusion~\cite{4} proposed a TSDF-based real-time scene reconstruction framework, integrating SEEM for 2D instance segmentation and Hungarian matching for instance tracking. While effective, OpenFusion relies on static semantic initialization, where instance embeddings are determined from the first observation and remain unchanged. This can lead to semantic drift, where partial observations cause incorrect labels to persist over time. Additionally, OpenFusion’s attribute regression mechanism struggles with complex, spatially dependent queries, such as “the chair near the window.”

Building on OpenFusion, we propose OpenFusion++, a real-time open-vocabulary scene understanding system that enhances semantic consistency, instance boundary precision, and query responsiveness. Our approach introduces confidence-guided point sampling to refine 3D instance boundaries, an adaptive semantic cache to mitigate drift through multi-view feature fusion, and a dual-path query framework that combines object-centric and spatial-aware features for precise retrieval of similar instances.

\begin{figure}[t] 
  \centering
  
  \begin{subfigure}{0.266\textwidth}
    \includegraphics[width=\linewidth]{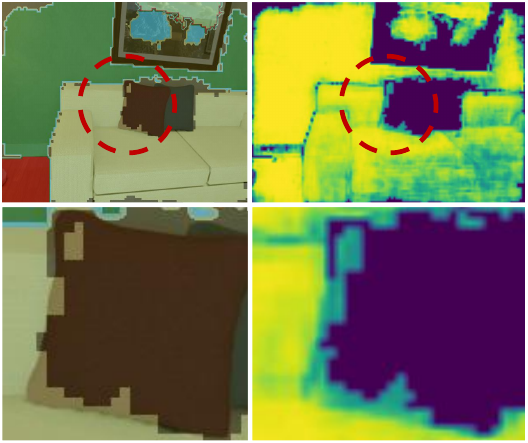}
    \caption{}
    \label{fig:subfig_a}
  \end{subfigure}
  \hfill
  \begin{subfigure}{0.2135\textwidth}
    \includegraphics[width=\linewidth]{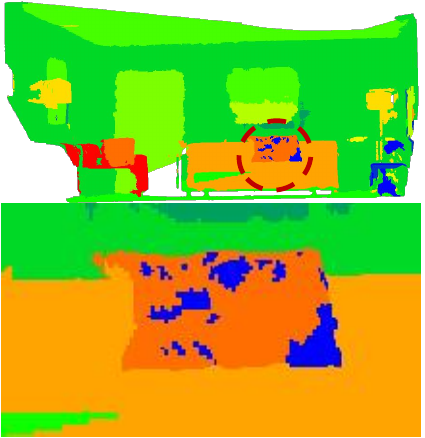}
    \caption{}
    \label{fig:subfig_b}
  \end{subfigure}
  \caption{Semantic confusion problem. The left side of (a) presents segmentation results, where different colors represent different instances, while the right side displays the corresponding confidence maps. Darker regions indicate lower confidence, with black representing the background class. A zoomed-in view at the bottom highlights local details. (b) demonstrates a semantic confusion issue where the pillow is partially eroded by the sofa.} 
  \label{sample_issue}
\end{figure}

\section{Methodology}
As shown in Fig.~\ref{overview}, our system processes RGBD streams for instance segmentation and region-level semantic feature extraction. OpenFusion refines the semantic-geometric map using TSDF for geometric mapping and Hungarian matching for instance registration. Building on this, our enhancements focus on global geometry, semantic updates, and query understanding. Confidence-based point sampling sharpens instance boundaries, while a semantic caching module preserves the best global semantic embedding. A two-branch query framework, integrated with multiple VLFMs, mitigates modality differences between queries and semantic maps, improving target instance retrieval. The method section first reviews key technologies of baseline before detailing our approach, module by module.

\subsection{Problem Definition and Foundations}

The goal of real-time open-vocabulary scene understanding is to incrementally build a structured 3D scene representation that integrates geometric reconstruction, instance-level segmentation, and class-agnostic semantic embeddings from continuous RGB-D inputs \(\{I_t, D_t\}_{t=1}^T\). Our approach relies on key principles of volumetric mapping, and semantic fusion to achieve accurate and scalable scene understanding.

The system utilizes the TSDF for real-time geometric reconstruction. The 3D space is discretized into dynamically allocated voxel blocks \( B_i \), where each block represents a localized region of the scene. Each voxel block stores a TSDF value representing the signed distance to the nearest surface, truncated within a predefined range. Additionally, it maintains color attributes, which are weighted RGB values aggregated from multi-view observations.  

For open-vocabulary semantic understanding, the system extracts 2D instance masks \(\{M_t^k\}_{k=1}^K\) and their semantic embeddings \(\{\mathbf{e}_t^k \in \mathbb{R}^d\}_{k=1}^K\) per frame using the SEEM model. Then it projects global 3D instances \(\mathcal{G} = \{G_j\}_{j=1}^J\) onto image space at current frame $I_j$ computes a many-to-many  Intersection over Union (IoU) cost matrix with the current frame's segmentation, and registers instances via optimal one-to-one linear assignment using Hungarian matching. Then, back-projecting matched masks into 3D space using depth map \(D_t\), generating instance point clouds \(P_j \subseteq B_i\). In the next step, we link voxel blocks \(B_i\) to instance ID \(j\), storing coordinates \(\mathbf{x}_p \in \mathbb{R}^3\) and per-point confidence \(c_p \in [0,1]\). The point density is controlled and instances are updated by independent point sampling in localized areas \(B_i\).

The global scene map adopts a hierarchical structure. Each voxel block \(B_i = (\mathbf{x}_p, c_p, j)\) is linked to an instance \(j \in \mathcal{G}\), where \(\mathbf{x}_p \in \mathbb{R}^3\) represents the 3D coordinates, and \(c_p \in [0,1]\) denotes the per-point confidence. Additionally, the system maintains a global semantic embedding set:  

\[
S_{\text{global}} = \{(j, \mathbf{e}_j) \mid j \in \mathcal{G} \},\eqno{(1)}
\]
where \(\mathbf{e}_j\) is fixed based on initial observations.

\begin{figure}[t]
  \centering
  \includegraphics[width=0.48\textwidth]{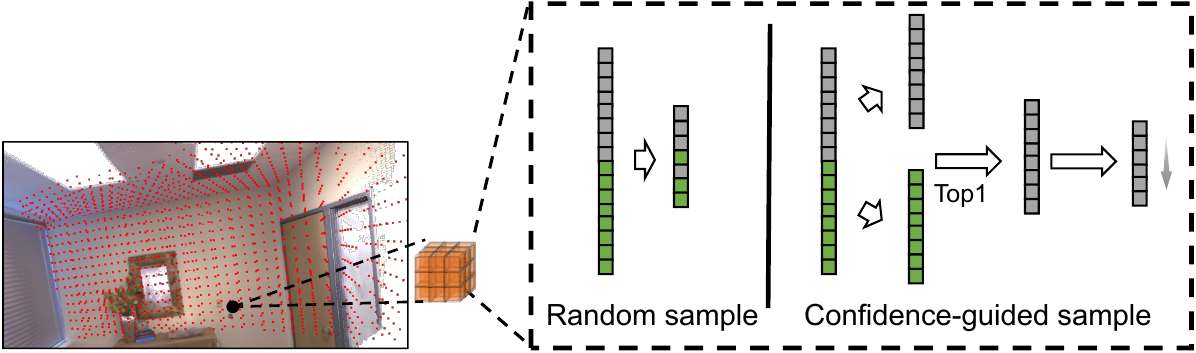}
  \caption{Sampling strategies within voxel blocks. The figure compares random sampling and confidence-guided sampling within a single voxel block. Different colored squares represent 3D points from different instances.}
  \label{conf_sample}
\end{figure}

\subsection{Confidence-guided Point Sampling}
\label{sec1}

After performing instance tracking across frames using the linear assignment, the system back-projects the 2D instance mask $M_t^k$ into 3D space using the depth map $D_t$, generating an instance point cloud and registering it to the corresponding voxel block $B_i$. Due to the high density of depth scans, each block maintains a fixed-capacity 3D point set of size $N$. When instance registration is completed, we sample to complete the screening of 3D points in each voxel block.

When 2D instance masks have precise segmentation boundaries, instance conflicts rarely occur within a voxel block. However, we observe that SEEM, which directly references object-level masks, often produces imprecise instance boundaries. As shown in Fig.~\ref{sample_issue}, the sofa instance encroaches on the pillow boundary, leading to semantic conflicts (Fig.~\ref{fig:subfig_b}). By analyzing the pixel-wise confidence map $C_t^k \in [0,1]^{H \times W}$ generated by SEEM, we find that eroded boundaries exhibit lower confidence values, providing a key insight for optimizing 3D instance representation.

Based on this observation, we propose a confidence-guided point sampling strategy. As shown in Fig. \ref{conf_sample},  given that voxel block sizes are typically at the centimeter scale, we assume that each block contains only one instance. For each candidate instance point set $P_{ij}$ within voxel block $B_i$, we compute its average confidence as:

$$
\bar{C}_{ij} = \frac{1}{|P_{ij}|} \sum_{p \in P_{ij}} C_t^k(p).\eqno{(2)}
$$

Following the geometric prior that a centimeter-scale voxel block contains a single instance, we select the point set with the highest $\bar{C}_{ij}$, 
$$
P_i^* = \arg\max_j \bar{C}_{ij},\eqno{(3)}
$$
and assign its corresponding semantic label as the latest instance representation for the region.

This process is executed incrementally during mapping, utilizing sequence data to refine 3D instance representation. Compared to traditional random sampling, our method of explicit selection preserves the best geometric representation, effectively avoiding instance conflicts.

\begin{figure}[t]
  \centering
  \includegraphics[width=0.48\textwidth]{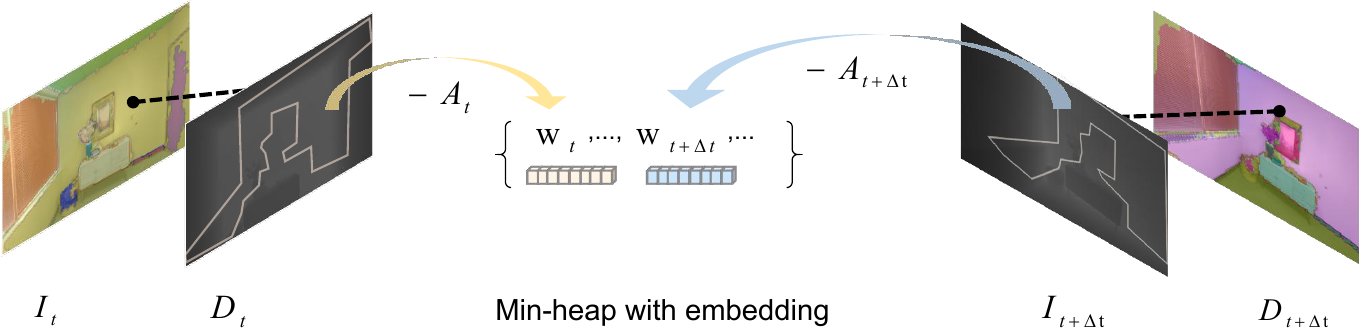}
  \caption{Semantic cache. The figure shows the process of dynamically inserting semantic embeddings. The weights of semantic embeddings are managed by inverting the physical coverage area as a minimum heap.}
  \label{cache}
\end{figure}
\subsection{Adaptive Semantic Cache}
\label{sec2}

Confidence-guided instance refinement improves boundary geometric accuracy, but instance semantic embedding remains limited by the locality of single-view observations. The incremental mapping process involves multi-view observations of instances.  For SEEM or CLIP, correctly regressing semantics from partial instances becomes particularly challenging when distinctive features are lacking. If the initial observation is directly used as the overall instance semantics, it may lead to semantic drift. Therefore,  it is crucial to integrate multi-view data to enhance global semantic representation.

To address this issue, we propose a global update method of semantic features. As shown in Fig.~\ref{cache}, we achieve robust optimization of global semantics through incremental fusion of multi-view features. Specifically, we use observation completeness as an evaluation metric, utilizing the actual physical area of each 2D instance as a filtering criterion. For a 2D instance mask \( M_t^k \), its physical coverage area \( A_t^k \) is computed via pixel-wise projection:

$$
A_t^k = \sum_{p \in M_t^k} \frac{f_xf_y}{D_t(p)^2},\eqno{(4)}
$$
where \( f_x, f_y \) are the camera focal lengths, and \( D_t(p) \) represents the depth value at pixel \( p \). 

This formula accumulates the physical projection area of all pixels within the mask, precisely quantifying instance observation completeness. As the viewpoint transitions from local to global, \( A_t^k \) monotonically increases, providing a dynamic weighting basis for semantic filtering and fusion.

Based on this, we propose using a hash-guided min-heap as a semantic cache, which dynamically sorts semantic embeddings in ascending order according to the inserted instance area. To ensure the accuracy of instance descriptions, for each instance \( G_j \), we maintain a min-heap \( H \) of size \( N \), where historical semantic embeddings are stored in ascending order based on the weight \( W = -A_t^k \). When a new observation \( \mathbf{e}_j^{\text{new}} \) is inserted, if the heap is full, the smallest area entry is removed.

Global semantic embeddings are updated through area-weighted multi-view feature fusion:

$$
\mathbf{e}_j \leftarrow \frac{\sum_{i=1}^{N} A_j^i \cdot \mathbf{e}_j^i}{\sum_{i=1}^{N} A_j^i + \epsilon},\eqno{(5)}
$$
where \( A_j^i \) represents the area weight of the \( i \)-th observation, $\epsilon=1e-8$ for numerical stability. This design preserves feature diversity and prevents information loss caused by normalization.

Furthermore, we construct a hash dictionary based on instance IDs to maintain the unique global semantics  \( S_{\text{global}} \) of each instance. Whenever a new observation is successfully inserted into the cache, a semantic fusion process is triggered to update the global semantics, improving response accuracy during real-time queries.

\begin{figure}[t]
  \centering
  \includegraphics[width=0.48\textwidth]{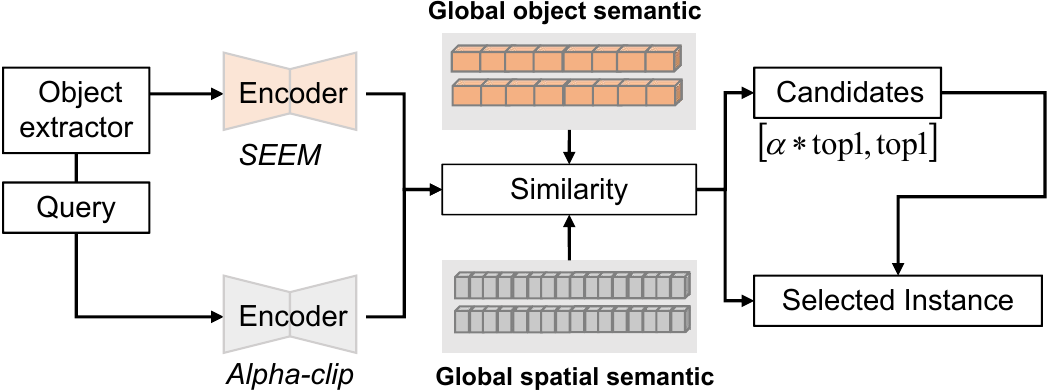}
  \caption{Hierarchical retrieval framework. The query in the figure goes through an object extractor to get the semantic embedding of the query object, and the instance features from the image in the same feature space with the incremental mapping process for similarity computation, and the range of candidate instances is controlled by the parameter alpha. Alpha-clip matches the query's overall encoding with the instance environment features and filters the best instances within the candidate range.}
  \label{retrieval}
\end{figure}
\subsection{Object-to-Spatial Query Method}
\label{sec3}

SEEM extracts object features linked to attributes, enabling simple category queries but struggling with specific instance queries based on attributes like shape, category, and color. Nested querying, such as "the cabinet near the door," is often used, but SEEM’s text encoder is limited to category sensitivity and cannot handle complex queries. While CLIP-based models can describe environments, they struggle with fine-grained distinctions between overlapping categories (e.g., "wall" vs. "door"). This limitation arises because multi-modal contrastive learning models like CLIP learn image and text representations in a shared semantic space, making them unsuitable for direct object-level queries.

To respond to complex queries, we design a two-stage method (Fig. \ref{retrieval}). First, the system extracts core object semantics from the query using the \texttt{en\_core\_web\_sm} model from spaCy (e.g., "cabinet") and the SEEM text encoder to generate an object embedding vector: $\mathbf{e}_{\text{obj}} \in \mathbb{R}^{d}$. Next, we compute the cosine similarity between this embedding and all instances in \( S_{\text{global}} \), 

$$
s_j = \cos(\mathbf{e}_{\text{obj}}, \mathbf{e}_j),\eqno{(6)}
$$
which quantifies the relevance to the target category.

To adapt dynamically to different query ambiguities, the system does not apply a fixed Top-K strategy. Instead, it filters candidate instances based on a threshold parameter \( \alpha \in (0,1) \). First, we determine the highest similarity score:

$$
s_{\text{top1}} = \max_j s_j\eqno{(7)}
$$
Then, the candidate set is constructed as:

$$
C_{\text{cand}} = \{ G_j \mid s_j \in [\alpha \cdot s_{\text{top1}}, s_{\text{top1}}] \}\eqno{(8)}
$$

Alpha-CLIP~\cite{25} is a variant of CLIP that focuses on regional features and extracts contextual information centered on the object based on a given mask. It enhances the richness of instance semantics during mapping. While in the query session, this branch aims to fully understand the query. In the set of candidate instances screened by the SEEM branch, we use the same similarity computation process and then choose the instance with the highest score as the response result. The combination of SEEM’s accurate category responses with CLIP’s environmental understanding enables precise interpretation of complex queries. This helps differentiate the target among similar instances, significantly improving the scene understanding system’s response capabilities.

\section{Experiments}

In this section, we conduct a comprehensive qualitative and quantitative evaluation of the proposed open-vocabulary scene understanding system on the ICL~\cite{28}, ScanNet~\cite{27}, ScanNet++~\cite{26}, and Replica~\cite{29} datasets. Our focus is on semantic segmentation of scenes, comprehension of complex queries, and response accuracy. For this reason, we additionally performed two ablation studies to validate the effectiveness of the proposed approaches.

All experiments were conducted on an NVIDIA GeForce RTX 4060 Laptop GPU (8 GB VRAM). The voxel map follows OpenFusion, maintaining \( N=16 \) points per voxel block and each block with a size in $8/512$ $m$. The cache stores up to three instance embeddings per object, and the threshold of query candidate range  \( \alpha \) is set to 0.8.

We evaluate the proposed method on ScanNet, ScanNet++, and Replica. ScanNet is a large-scale RGB-D video dataset designed for indoor scene understanding, containing 1,513 real-world indoor scans with approximately 2.5 million RGB-D frames. Each scan is annotated with instance-level semantic segmentation across 20 categories. ScanNet++ is an extension of ScanNet that expands the number of categories to 200 and increases both scene complexity and semantic segmentation granularity. Replica is a dataset providing high-fidelity, photorealistic 3D indoor scene models.

We select 26 sequences (8 of Replica, 10 of Scannet, and 8 of Scannet++) from the three datasets for rigorous evaluation of our method. The semantic segmentation results are obtained by querying the sequence-provided category list. To assess segmentation accuracy, we adopt two standard metrics: mean Accuracy (mAcc), and the frequency-weighted Mean Intersection over Union (f-mIoU).

\begin{table}[t]
    \centering
    \renewcommand{\arraystretch}{1.2}
    \caption{Comparison of the proposed method with the baseline on the Replica, ScanNet, and ScanNet++ datasets.}
    \resizebox{\linewidth}{!}{
    \begin{tabular}{l l c c}
        \hline
        \textbf{Dataset} & \textbf{Method} & \textbf{mACC (\%)} & \textbf{F-mIOU (\%)} \\
        \hline
        \multirow{2}{*}{Replica (8 seq)} & Openfusion & 38.21 & 60.07 \\
        & Openfusion++ (Ours) & \textbf{44.50} & \textbf{62.97} \\
        \hline
        \multirow{2}{*}{Scannet (10 seq)} & Openfusion & 62.46 & 65.80 \\
        & Openfusion++ (Ours) & \textbf{64.40} & \textbf{67.62} \\
        \hline
        \multirow{2}{*}{Scannet++ (8 seq)} & Openfusion & 27.04 & 45.31 \\
        & Openfusion++ (Ours) & \textbf{31.68} & \textbf{47.21} \\
        \hline
    \end{tabular}
    }
    \label{tab:method_comparison}
\end{table}

\subsection{Semantic Segmentation Performance Experiments}

As shown in Table~\ref{tab:method_comparison}, the proposed global 3D point cloud sampling and semantic update strategy demonstrate significant advantages in the semantic segmentation tasks on ScanNet, ScanNet++, and Replica datasets. Compared to the OpenFusion baseline, our method achieves an average 4.2\% improvement in mAcc across the three datasets, with the most notable gain observed on Replica (+6\%), which involves fine-grained classification tasks.
\begin{figure*}[thpb]
  \centering
  \includegraphics[width=1.0\textwidth]{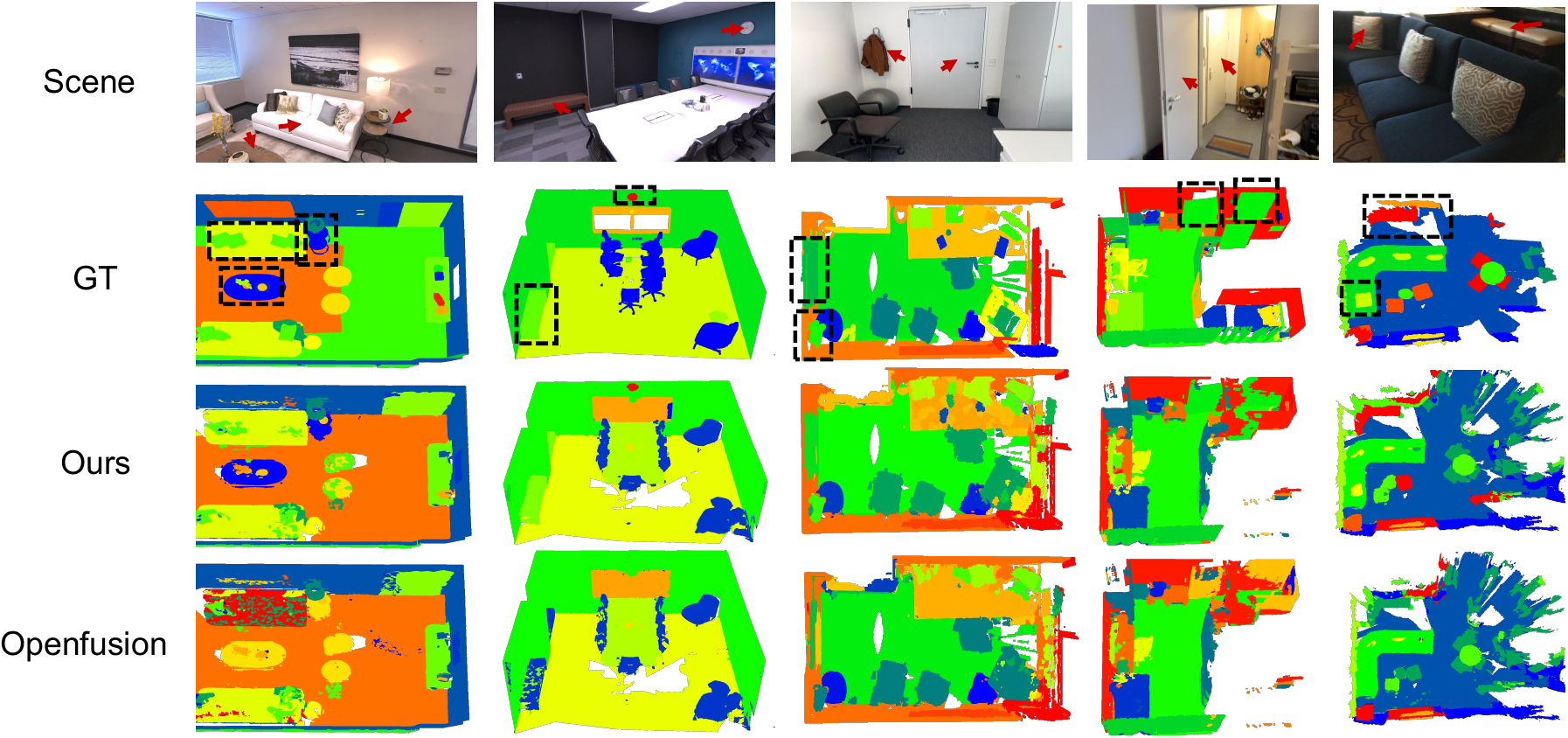}
  \caption{Qualitative experimental results. Based on the list of categories corresponding to each scenario, we obtain the semantic labels of the instances through a query framework, where different labels are separated by color differences. The top image is the scenario, with the red arrows pointing to the instances where our approach has a significant improvement compared to the baseline. The GT layer provides a reference for the real labels of the instances.}
  \label{metric}
\end{figure*}

\begin{figure*}[thpb]
  \centering
  \includegraphics[width=1.0\textwidth]{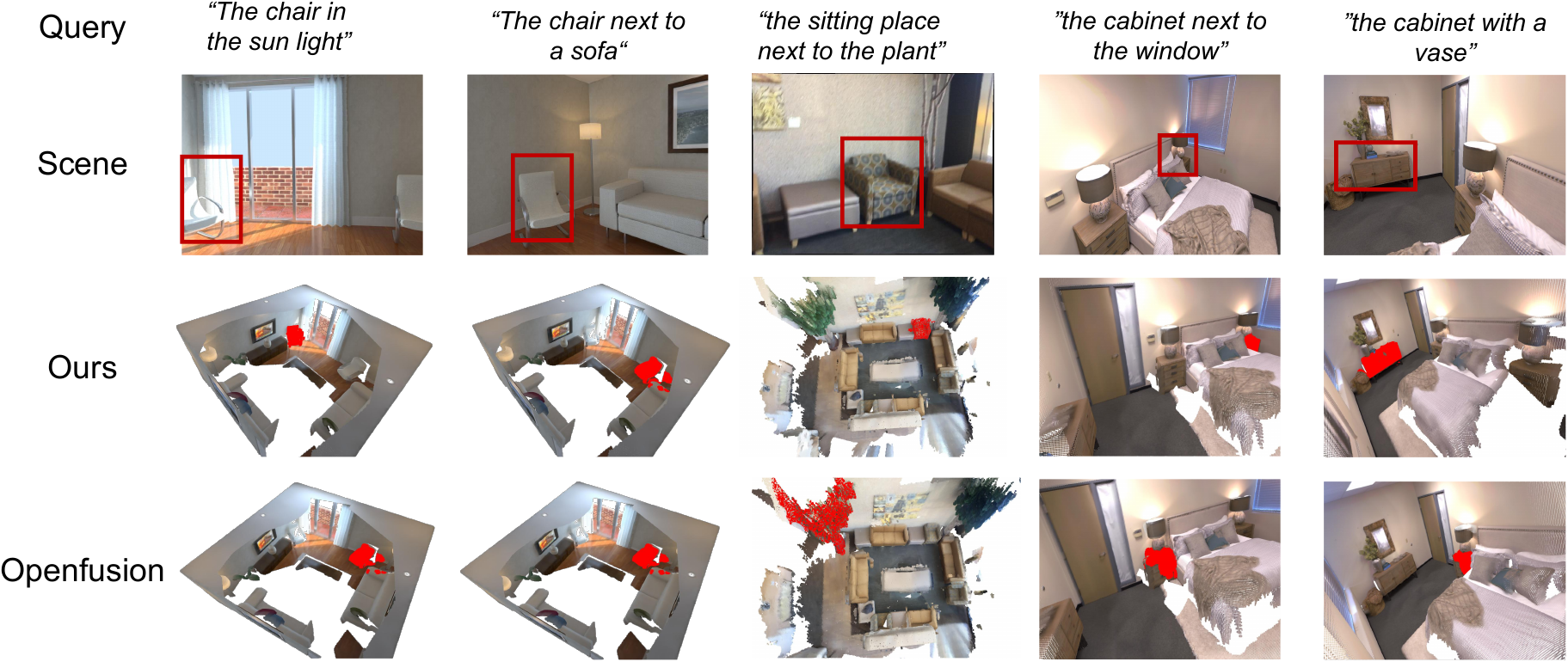}
  \caption{Visualisation of query results. The figure shows the results of the response given by the system for a given nested query. The top layer shows the given input and the target object in the scene is highlighted using boxes.}
  \label{query}
\end{figure*}

The qualitative results (Fig.~\ref{metric}) further validate the effectiveness of our method. In Replica and ScanNet scenes, our approach produces clear and coherent semantic boundaries, whereas the baseline method exhibits semantic fragmentation within instances due to random sampling strategies (e.g., sofa regions mixed with pillow labels). Moreover, cross-instance semantic confusion arises in the baseline due to the lack of global semantic updating.



These experimental results indicate that our global optimization strategy effectively suppresses semantic drift, enhancing fine-grained understanding in complex scenes.

\subsection{Query Response Experiments}
We conducted complex query qualitative tests on the ICL, Replica, and ScanNet datasets to evaluate the two-stage query framework's ability to distinguish similar instances.

As illustrated in Fig.~\ref{query}, in indoor scenes containing multiple chairs, OpenFusion relies on single-view semantic embeddings, leading to repetitive retrieval of the same instance (e.g., the chair with the highest observation quality), while ignoring other similar instances.

Our approach accurately selects the target from semantically similar candidates by parsing environmental context (e.g., adjacent furniture layout and spatial relationships).

These experimental findings demonstrate that the hierarchical query mechanism, by integrating object attributes with environmental semantics, significantly improves the robustness of complex queries, providing reliable support for robotic object search and navigation tasks.

\begin{figure*}[t]
  \centering
  \includegraphics[width=1\textwidth]{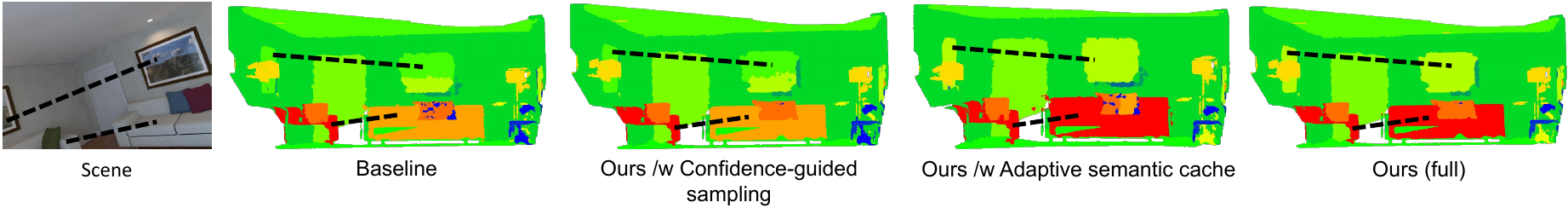}
  \caption{Ablation studies of map update strategies. The black dashed line connects multiple instances with the same semantics.}
  \label{ablation_mertric}
\end{figure*}

\subsection{Ablation Studies}
To validate the independent contributions of each module, we performed a systematic ablation study on the Replica and ICL datasets. As shown in Table~\ref{tab:replica_ablation}, in the 8 test sequences of Replica, introducing the confidence-guided sampling alone increased instance-level semantic mAcc by 3.73\%, enhancing instance boundaries, but decreased f-mIoU by 1.87\%. 

This was caused by the confidence filtering strategy slightly expanding the semantic coverage of some categories, like confusing doorframe edges with walls, while still reducing boundary noise. Adding the global semantic update module improved both mAcc and f-mIoU by 6.29\% and 2.90\%, respectively, as shown in Fig. \ref{ablation_mertric}, significantly reducing semantic ambiguity through multi-view feature fusion (e.g., recovering wall paintings' complete semantics).

\begin{figure}[t]
  \centering
  \includegraphics[width=0.48\textwidth]{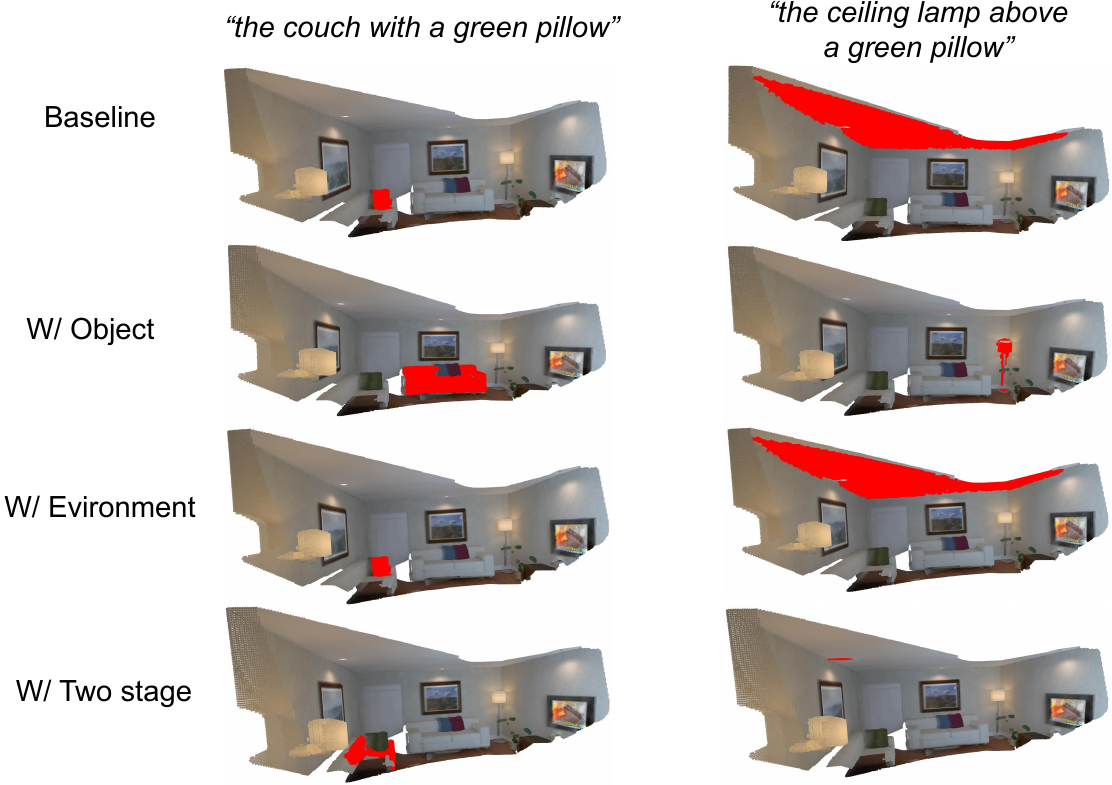}
  \caption{Results of query ablation experiments. Two types of queries are designed for similar instances in the figure. 'Object' denotes a query based on the core object through the SEEM branch, while 'Environment' represents a query through the alpha-clip branch, and "Two-stage" corresponds to a query that passes through the full hierarchical framework. The query results are marked in red.}
  \label{ablation_query}
\end{figure}

For the hierarchical query framework, a progressive ablation on the ICL dataset (Fig.~\ref{ablation_query}) showed that relying solely on SEEM object center encoding yielded high-quality instances but missed similar targets. Using only CLIP full-text encoding led to random incorrect results due to text-instance alignment biases. In contrast, the complete two-stage method (object selection and environmental matching) accurately located target instances, demonstrating the necessity of object-environment collaborative reasoning.

\begin{table}[t]
    \centering
    \renewcommand{\arraystretch}{1.2}
    \caption{Ablation Study of Map Update Strategies.}
    \resizebox{\linewidth}{!}{ 
        \begin{tabular}{lcc}
            \hline
            & \textbf{mACC (\%)} & \textbf{F-mIOU (\%)}\\
            \hline
            Baseline & 38.21  & 60.07 \\
            Ours w/ Confidence-guided sampling & 41.94 (+3.73) & 58.20 (-1.87)  \\
            Ours w/ Adaptive semantic cache & 40.50 (+2.29)& 61.17 (+1.10) \\
            Ours (full) & \textbf{44.50 (+6.29)}& \textbf{62.97 (+2.90)} \\
            \hline
        \end{tabular}
    }
    \label{tab:replica_ablation}
\end{table}

\section{Conclusion}
This paper introduces OpenFusion++, a real-time open-vocabulary scene understanding system that addresses key issues in existing methods, such as blurry instance segmentation boundaries, insufficient global semantic updates, and limited complex query response capabilities. By utilizing confidence-guided 3D point sampling, an adaptive semantic caching mechanism, and a dual-path feature query framework, our approach significantly enhances performance. The experiment results demonstrate the system's ability to maintain clear semantic boundaries in complex scenes and accurately respond to environment-related queries.

The main limitation of this method is its reliance on SEEM's 2D segmentation quality, which may degrade under real-world challenges like motion blur.  Additionally, during incremental processing, voxel block creation may occasionally assign incorrect instances or labels due to confidence sampling. While random sampling allows subsequent tracking to correct errors, confidence sampling dominates insertions, preventing correction. In the future, this issue could be mitigated by adopting batch insertion with weighted confidence sampling instead of frame-level operations.

\bibliographystyle{IEEEtran}
\bibliography{ref}

\end{document}